\documentclass{article}
\usepackage[ruled]{algorithm2e}
\usepackage[accepted]{mlsys2024_for_arxiv}
\usepackage{microtype} 
\usepackage{setspace}
\usepackage{hyperref}       % hyperlinks

\usepackage{subcaption}
\usepackage[utf8]{inputenc} % allow utf-8 input
\usepackage[T1]{fontenc}    % use 8-bit T1 fonts
\usepackage{url}            % simple URL typesetting
\usepackage{booktabs}       % professional-quality tables
\usepackage{amsfonts}       % blackboard math symbols
\usepackage{nicefrac}       % compact symbols for 1/2, etc.
\usepackage{xcolor}         % colors
\usepackage{multirow}

\usepackage{amsmath}
\usepackage{amssymb}
\usepackage{mathtools}
\usepackage{amsthm}

\usepackage[textsize=tiny]{todonotes}

\usepackage{enumitem}

\newcommand{\name}{\small FAIR \normalsize}
\newcommand{\robe}{\small ROBE\normalsize}
\newcommand{\roast}{\small ROAST\normalsize}
\newcommand{\heterofl}{\small HETEROFL\normalsize}
\newcommand{\fedhm}{\small FEDHM\normalsize}
\newcommand{\fair}{\small FAIR\normalsize}
\newcommand{\fairhom}{\small FAIR{-}HOM\normalsize}
\newcommand{\fairhet}{\small FAIR{-}HET\normalsize}
\newcommand{\fulltrunc}{\small FULL{-}TRN\normalsize}

\newcommand{\ssection}[1]{\section{#1}}
\newtheorem{theorem}{Theorem}[section]

\newtheorem{lemma}[theorem]{Lemma}
\newtheorem{corollary}[theorem]{Corollary}
\theoremstyle{definition}
\newtheorem{definition}[theorem]{Definition}

\theoremstyle{remark}

\mlsystitlerunning{FAIR: Federated Averaging in Random Subspaces}

\begin{document}

\twocolumn[
\mlsystitle{Heterogeneous federated collaborative filtering using FAIR: Federated Averaging in Random Subspaces}

% unless you've provided the [accepted] option to the mlsys2024


\mlsyssetsymbol{equal}{*}

\begin{mlsysauthorlist}
\mlsysauthor{Aditya Desai}{rice}
\mlsysauthor{Benjamin Meisburger}{rice}
\mlsysauthor{Zichang Liu}{rice}
\mlsysauthor{Anshumali Shrivastava}{rice,tai}
\end{mlsysauthorlist}

\mlsysaffiliation{rice}{Department of Computer Science, Rice University, Houston, Texas}
\mlsysaffiliation{tai}{ThirdAI Corp, Houston, Texas, USA}

\mlsyscorrespondingauthor{Aditya Desai}{apd10@rice.edu}
\mlsyskeywords{Machine Learning, MLSys}

\vskip 0.3in

\begin{abstract}
Recommendation systems (RS) for items (e.g., movies, books) and ads are widely used to tailor content to users on various internet platforms. Traditionally, recommendation models are trained on a central server. However, due to rising concerns for data privacy and regulations like the GDPR, federated learning is an increasingly popular paradigm in which data never leaves the client device. Applying federated learning to recommendation models is non-trivial due to large embedding tables, which often exceed the memory constraints of most user devices. To include data from all devices in federated learning, we must enable collective training of embedding tables on devices with heterogeneous memory capacities. Current solutions to heterogeneous federated learning can only accommodate a small range of capacities and thus limit the number of devices that can participate in training. We present Federated Averaging in Random subspaces (\fair), which allows arbitrary compression of embedding tables based on device capacity and ensures the participation of all devices in training. {\fair} uses what we call \textit{consistent} and \textit{collapsible} subspaces defined by hashing-based random projections to jointly train large embedding tables while using varying amounts of compression on user devices. We evaluate {\fair} on Neural Collaborative Filtering tasks with multiple datasets and verify that {\fair} can gather and share information from a wide range of devices with varying capacities, allowing for seamless collaboration. We prove the convergence of {\fair} in the homogeneous setting with non-i.i.d data distribution. Our code is open source at {https://github.com/apd10/FLCF}%{https://anonymous.4open.science/r/FAIR\_FederatedCollaborativeFiltering}.
\end{abstract}
]
\printAffiliationsAndNotice{}
\ssection{Introduction}
% Recommender systems and their importance.
Recommendation systems (RS) form the backbone of a good user experience on various platforms, such as e-commerce websites, social media, streaming services, and more. RS solves the issue of information overload by helping users discover the information most pertinent to them. To provide personalized suggestions, RS collects user features such as gender, age, geographical information, etc., and past user activity on the platform. User feedback can be categorized as either explicit feedback, such as ratings, or implicit feedback, such as views, clicks, or time spent on an item.

% why federated learning in collaborative filtering - privacy, etc., federated learning is the only solution.
Traditionally, the user feature and interaction data is collected on a server, and RS models are trained centrally. However, several studies \cite{calandrino2011you, lam2006you, mcsherry2009differentially} have exposed the privacy risk associated with the centralized collection of data. Even seemingly benign interaction data, such as movie ratings, can be used to deduce personal details such as age, gender, and political affiliations \cite{weinsberg2012blurme, aonghusa2016don, narayanan2008robust}. Owing to such risks, regularization of data usage and protection is increasingly necessary. One such legislation in the European Union is the General Data Protection Regulation (GDPR), which regulates data usage conditions by platforms. Federated learning, a distributed machine learning paradigm, ensures that user data never leaves its respective device---an apt solution for data privacy.
\begin{figure*}[t]
\centering
    \includegraphics[scale=0.3]{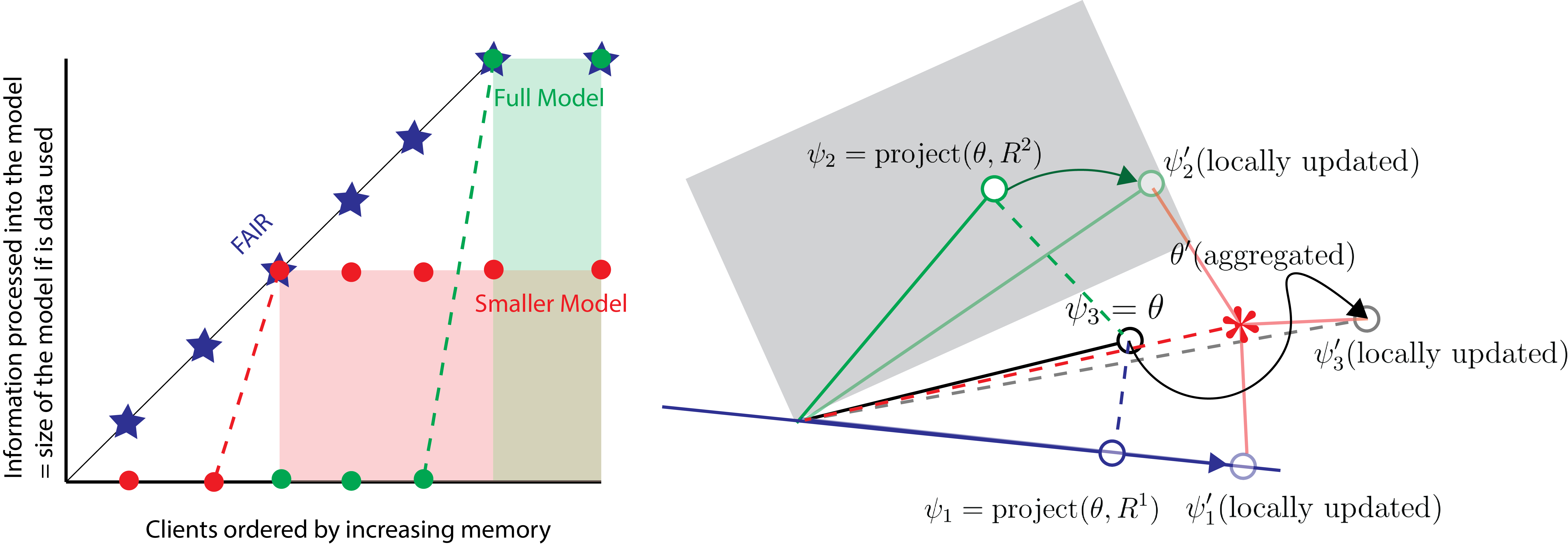}
    %\caption{Generally model size determines which devices can participate in the federated learning : i.e., depending on model size, device data is either completely Incorporated in model or is ignored. FAIR customizes the quality of  information processed into the final model based on the device capacity, thus offering much smoother trade-off between device capacity and information used from that device}
    \caption{\textbf{[Left] }Consider the information contributed by a device as the model-size trained with local data. Illustration shows 2/7 devices can host full model. Thus, under traditional settings, we can only run federated learning with those two devices. To include more devices we need to reduce the model size. With a smaller model we can include, perhaps, 5/7 devices but the information quality is lower for all (red plot). \fair{} provides smooth trade-off where each device can host a different sized model and contribute information based on its capacity. \textbf{[Right]} Illustration of \fair{}. Let the parameter be in $R^3$, There are three devices which can store parameter sizes of 1 (blue), 2 (green) and 3 (black) respectively. In each round, \fair{} projects the $\theta$ vector (dark black circle) to random sub-spaces of dimensions 1 and 2. Devices with capacity of 3 can directly use $\theta$. Models are optimized inside the sub-spaces locally, and then aggregated back at the server (red) in $R^3$.}
    \label{fig:fair-tradeoff}    
%\vspace{-0.6cm}
\end{figure*}

% Training and deploying recommendation models on client devices issues large embedding tables. Give some numerals. 
RS models, such as collaborative filtering (CF)  models like MF \cite{koren2009matrix}, NeuMF \cite{he2017neural}, etc., and ad recommendation click-through models, such as DLRM \cite{dlrm}, Deep Factorization Model \cite{guo2017deepfm}, etc. have embedding tables for items and other categorical features. These embeddings are typically very large due to large product catalogs and other categorical data. For instance, Pinterest has billions of pins for recommendation \cite{pal2020pinnersage}, and Amazon offers 100s of millions of products---leading to multiple gigabyte-sized embedding tables. To train a federated learning-based RS model, the model must necessarily fit onto client devices. Given the scale of these embedding tables, many client devices (such as mobile phones or low-memory laptops) simply cannot store the full model.

% heterogeneous collaboration is needed. Give a concrete example. 

Thus, the problem of client device heterogeneity (with respect to memory capacity) is especially pronounced in the context of RS models. Devices such as desktops and laptops can potentially hold full RS embedding tables, whereas devices such as mobile phones cannot. Two naive solutions to this issue are: (A) to discard data from devices that cannot fit the model (data loss) or (B) to learn a smaller model to incorporate all devices (model loss). There is a natural trade-off where model size determines the devices that can participate in federated learning. This trade-off is shown in Figure~\ref{fig:fair-tradeoff}. FAIR can customize the quality of information obtained from a device based on its capacity while building a full-sized model on the server. Thus, with FAIR, all devices with arbitrarily small memory capacity can still participate in federated learning.

\begin{figure*}[t]
\begin{subfigure}{0.30 \textwidth}
    \centering
    \includegraphics[scale=0.30]{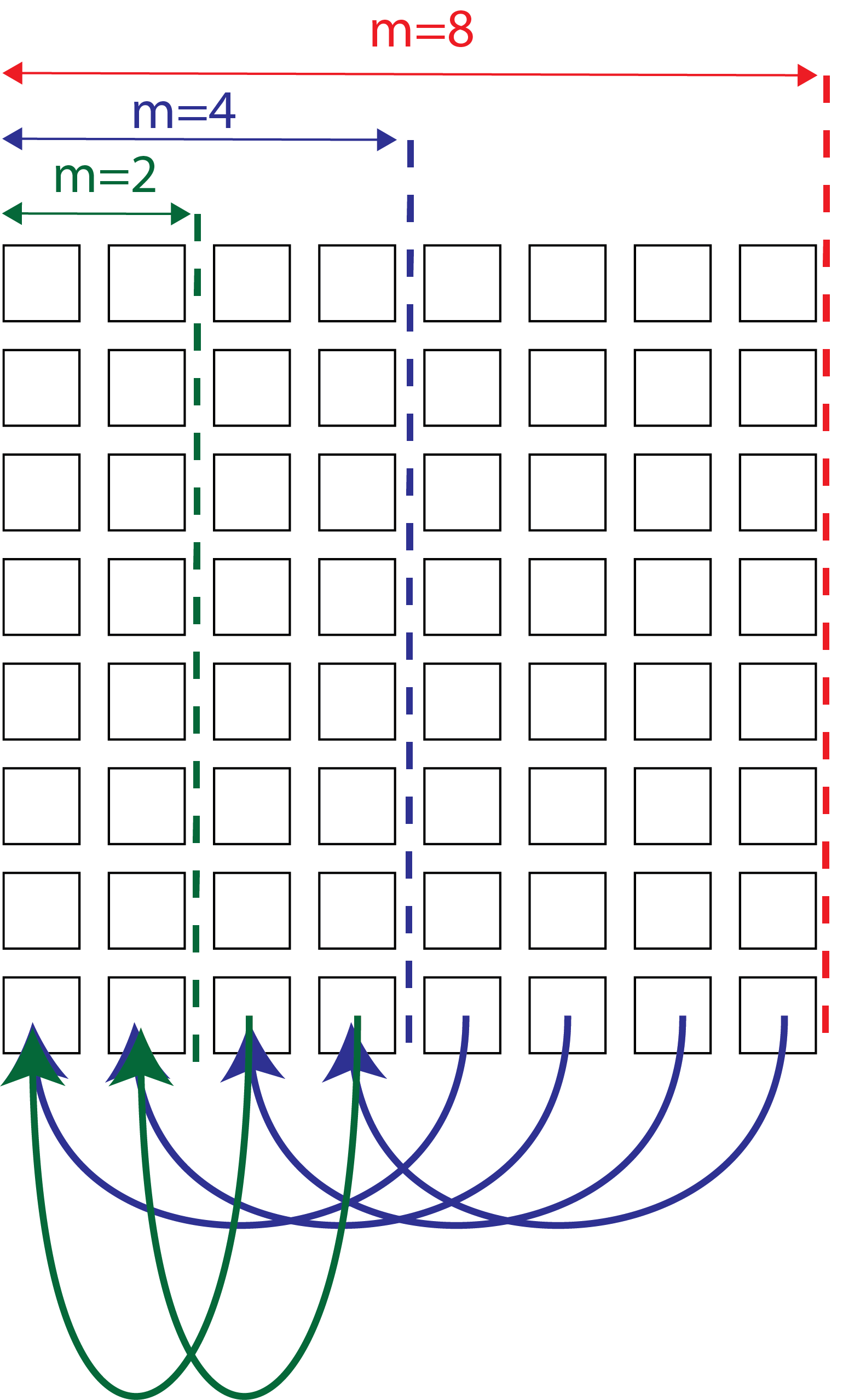}

\end{subfigure}
\hspace{0.5cm}
\begin{subfigure}{0.45\textwidth}
    \centering
    \includegraphics[scale=0.3]{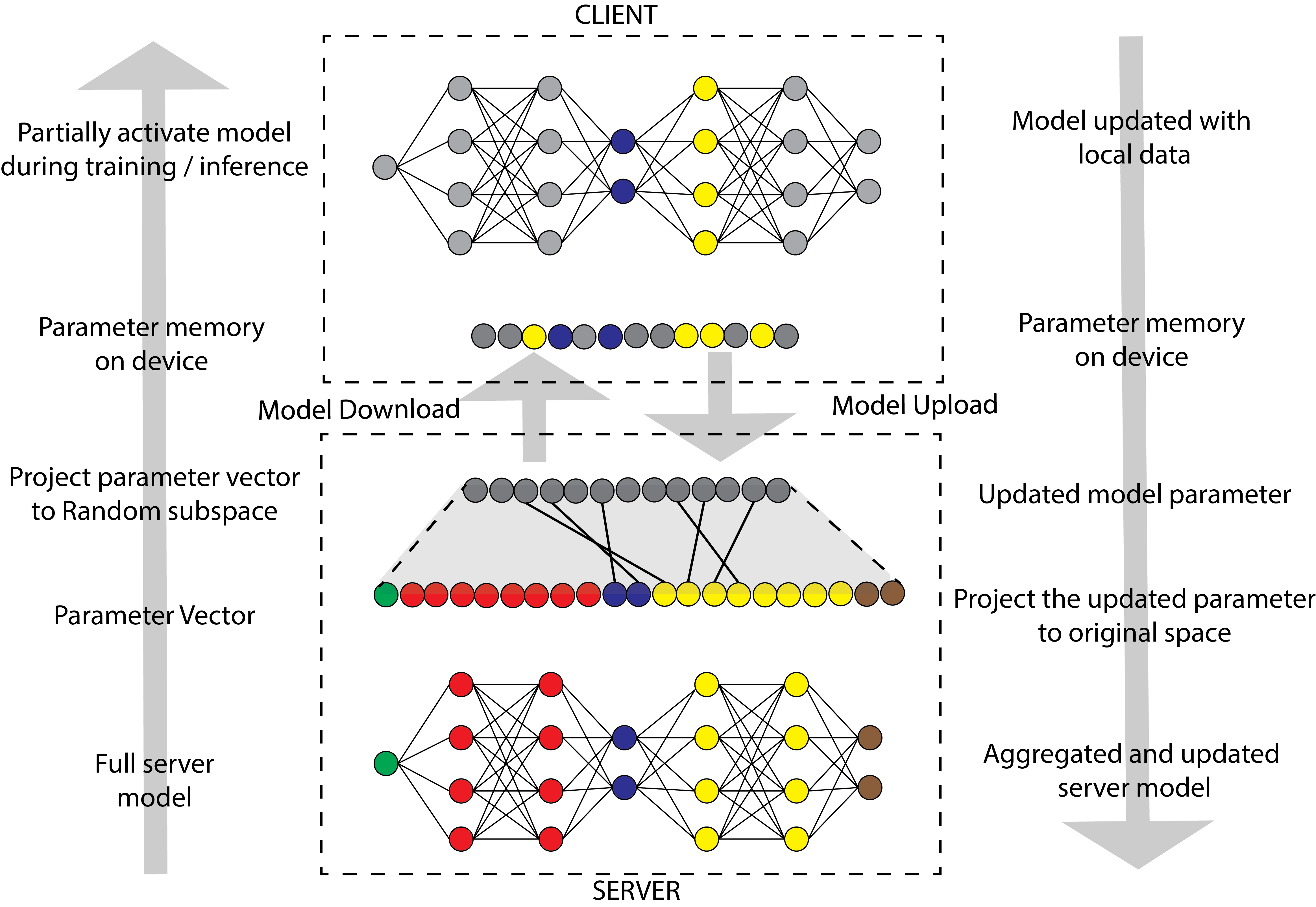} 
\end{subfigure}
\caption{\textbf{[Left]} Illustration of matrix clip-and-add to create a set of consistent and collapsible subspaces. The original $n\times m$ matrix is generated by the hash function $h$. Then depending on the target size $m_i$ for a device $i$, the matrix is cut into $m/m_i$ pieces and all the pieces are added to each other. \textbf{[Right]} A round associated with FAIR: The server projects its parameter vector $\theta$ into a smaller vector $\psi$ and sends this vector to the client. The client stores the smaller vector $\psi$ and trains it locally using memory efficient parameter sharing based kernels. The updated $\psi$ is sent back to the server which is expanded into $\theta$ and aggregated with parameters received from other devices. }  \label{fig:marixfolding} \label{fig:fair-process}
\end{figure*}

% FAIR summary
The general recipe of FAIR is shown in Figure~\ref{fig:fair-process}, which includes communicating with a device that cannot host the entire model. FAIR uses hashing-based random projections in combination with recent advances in efficient parameter-sharing kernels \cite{robe1,desai2022efficient} to reduce the memory footprint of the model and train on-device while only realizing a reduced memory cost. As illustrated in the toy example in Figure~\ref{fig:fair-tradeoff}, each device is associated with a random projection matrix (and hence a subspace). The parameters of the model are only stored in the subspace basis, thus reducing the model's memory footprint. Additionally, the parameters are optimized within this subspace and thus never fully realized in memory. Naively selecting random projections is detrimental to the collaborative training of the model. \fair{} proposes the use of \emph{consistent} and \emph{collapsible} subspaces which have provable convergence guarantees. A complete definition of consistent and collapsible subspaces is included in Section \ref{consistent-collapsible}. More details on intelligently building random projection matrices and model training without realizing the full model size are provided in Section \ref{sec:method}.

In our experimental results, we show that FAIR provides better quality models over both alternatives of data-loss (excluding devices) and model-loss (using smaller models). Additionally, it outperforms heterogeneous federated learning baselines such as FEDHM \cite{yao2021fedhm} and HETEROFL \cite{diao2020heterofl} on embedding tables. Furthermore, it enables us to include devices having smaller capacities that baseline methods cannot support. We also show that FAIR is a general strategy that can be extended to other model components, such as convolutions and MLPs.

\textbf{Limitations:} FAIR can be applied to embedding tables as well as other generic model components such as convolutions, MLPs, etc. However, one concern when applying FAIR to model components other than embedding tables is that FAIR does not reduce the computational workload of the compressed model. Thus, FAIR is most suitable for memory-heavy components such as embedding tables, which only require table lookups or indexing---making it a perfect solution to embedding-heavy recommendation models. We would, however, like to stress that FAIR can provide arbitrarily small memory footprints, which cannot be achieved by other methods, making it possible to include all client devices, which is impossible with existing approaches to heterogeneous federated learning.
\ssection{Related Work}
Federated learning is a distributed training approach in which model training occurs on user devices, preserving data privacy. A central server coordinates the process, sending initial model parameters to user devices, which subsequently train the model on local data and upload updated parameters to the server. Communication occurs in a staggered manner to reduce network overhead. Model updates from devices are aggregated using algorithms like FEDAVG \cite{mcmahan2017communication} or FEDProx \cite{li2020federated}, which average the model parameters from each participating device.

Research around federated learning in the context of recommendation systems has gained momentum in the past couple of years \cite{g1,g2,g3,g4,g5}. Some specific aspects of federated recommendation systems have been widely studied, such as privacy in \cite{p1,p2,p3,p4,p5,p6} and fairness \cite{f1,f2}. However, the memory issues associated with recommendation models--- primarily due to the enormous embedding tables part of most recommendation models---have not been investigated. This paper considers arguably the simplest setting of CF recommendation models to discuss the issue of large embeddings in federated learning.

Federated learning for CF models has been previously studied in the literature. Analysis has been made of existing algorithms using alternating least squares and stochastic gradient updates in the context of federated CF models \cite{ammad2019federated}. Differential privacy concerns have also been investigated within the context of communicated data in federated CF \cite{minto2021stronger,jiang2022fedncf}. To the best of our knowledge, we are the first to examine the issue of heterogeneous memory capacities of client devices, which need to participate in federated learning of a CF model and provide an efficient solution that can effectively collaborate information across heterogeneous models on different devices. 
 \begin{figure*}
    \centering
    \includegraphics[width=0.9\textwidth]{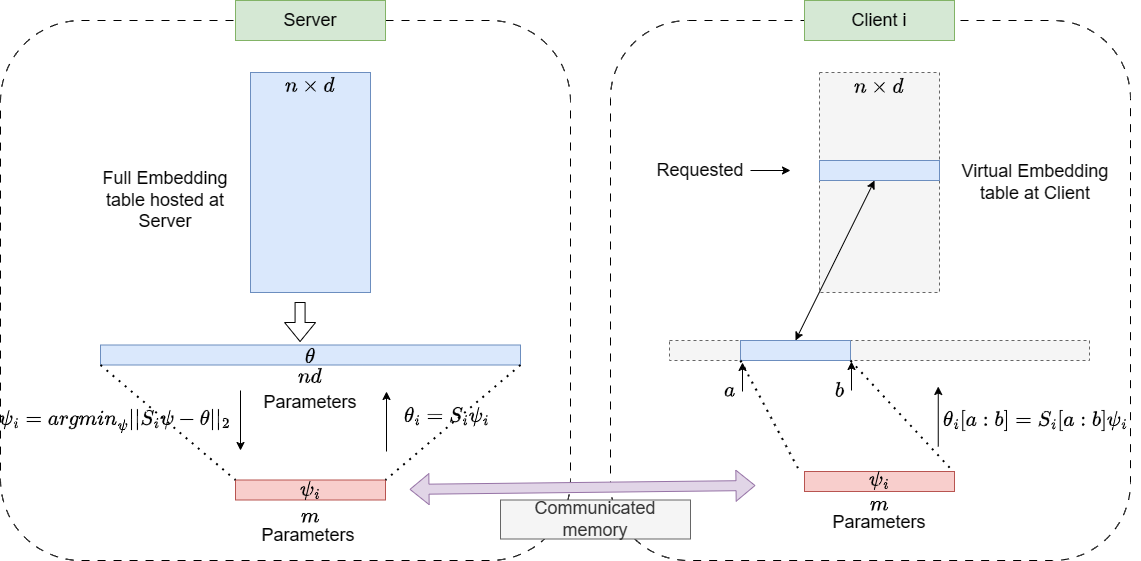}
    \caption{Illustration of how embedding tables are distributed to limited capacity devices, trained and updates are communicated taking one client as an example. In the figure, the entire embedding table of size $n\times d$ is hosted in server and has $nd$ sized parameter vector. The client device is associated with a (sparse) projection matrix $S$ generated with hash function. The server first projects the parameter vector to smaller column subspace of $S$ giving compressed vector $\psi$. The $\psi$ parameters are transmitted to device and trained on device. While training and inferring concrete parts of virtual embedding table (not realized) are recovered from the $\psi$ as required. While uploading, $\psi$ is uploaded to server and projected to server basis using $S\psi$. The server collects all such vectors from different clients. The $\theta_i$s thus received will then be used to update the server model $\theta$. In this paper we use FEDAVG, but we can potentially use any algorithm once all $\theta_i$s are collected}
    \label{fig:embedding_fair}
\end{figure*}
Recently, a few approaches have been explored for the problem of heterogeneity in storage and communication capacity in federated learning. The basic idea is to send smaller versions of models to clients with lower capacities. Both low-rank factorization and slices of weight matrices have been used to create smaller versions of models \cite{yao2021fedhm,diao2020heterofl}. However, in both approaches, there is a fundamental limit to the amount of memory reduction that can be achieved. The issue is especially burdensome when applying compression techniques for embedding tables (say $n\times d$), which, at times, are orders of magnitude larger than what user devices can store. While these methods can---at best---accommodate clients that need less than $d\times$ compression, FAIR can accommodate any client with an arbitrarily small memory capacity.

\begin{table}[t]
\small
\centering
\caption{Lowest capacity supported by various heterogenous FL methods}
\label{tab:memory}
\begin{tabular}{|c|c|c|c|}
\hline
($E : n \times d$) & FEDHM        & HETEROFL   & FAIR      \\ \hline
Max Compression & $< d \times$ & $d \times$ & Arbitrary \\ \hline
\end{tabular}
\vspace{-0.4cm}
\end{table}

\ssection{Background : Parameter sharing based compression}
Recently, randomized parameter sharing has shown promising results for model compression with HashedNet\cite{hashnet}, \robe{} \citep{robe1} and \roast{} \citep{roast}. \fair{} is also based on parameter sharing, and thus, we briefly explain how parameter sharing has been used for model compression in literature.

The general recipe of parameter sharing-based model compression is as follows. Consider a given model with a total number of parameters $n$. Under parameter sharing-based compression, there is a parameter repository $\mathcal{M}$ such that $|\mathcal{M}| < n$. The model weights are not explicitly stored in memory but are ``virtual". The only concrete parameter memory that is stored and trained is $\mathcal{M}$. The computational graph of the model is the same, and whenever we want to access the model weight, say a parameter $w$ of some layer inside the model, we use a hash function, say $h$, to locate the weight inside the repository of parameters. Thus, the value of $w$ is
\begin{equation}
    w = \mathcal{M}[h(id(w))]
\end{equation}
Optionally, another hash function $g$ with range $\{\pm1\}$ is multiplied with the retrieved value to obtain the final weight value. i.e. $w = g(id(w)) \mathcal{M}[h(id)]$. 

To the best of our knowledge, HashedNet was the first to propose parameter sharing-based compression. They used element-wise hash functions, which independently map each weight into the repository. One of the issues in this mapping is the cache-inefficiency as in order to retrieve, say, an embedding of size $d$, $d$ separate potentially faraway locations need to be accessed. The issue of cache-inefficiency is fixed in \robe{} for embedding tables and \roast{} for matrix multiplication operation.

\ssection{FAIR: Federated Averaging in Random Subspaces} \label{sec:method}
As mentioned earlier, \fair{} is best suited for models with large embedding tables, and thus, the focus of this paper is federated CF. However, the design of \fair{} is general and can be applied to other model components. In this section, we describe FAIR in the most general manner. We refer to the parameter vector $\theta$ (a flat vector of all model parameters being optimized), which may only include embedding table parameters in the context of CF models or other model components if needed.

% gives intuition, which can be maintained for sub-discussions.
The key idea in FAIR is that each device stores a model representation $\psi \in R^k$, which can be of smaller dimension than $\theta \in R^n$. The two representations are linked via a subspace defined by column space of $S \in R^{n\times k}$, also called the projection matrix. Simply put, in order to get $\psi \in R^k$, we project $\theta$ on the column subspace of $S$, and $\psi$ is the projected vector in the new basis of column-space ($S$). To recover the model in the server basis, we can just compute $S\psi$. When running the model on the client, we always maintain $\psi$ in memory but recover the parts of the model, say $\theta[a:b]$ for some $a,b$ in the original space using $S[[a:b], :] \psi$ as needed. This is illustrated in Figure~\ref{fig:fair-process} (right) and Figure~\ref{fig:embedding_fair}. We will discuss the details of these operations and how to choose these subspaces below.

\paragraph{Notation (federated learning):} We will use the following notation throughout the paper. The total number of devices is $N$, and in each round of federated learning, we will be sampling $K$ ($K \leq N$) devices. The training is conducted for $T$ rounds, each of which consists of $E$ steps of SGD. The optimization function associated with each device and its data is denoted as $F_i(\theta, \mathcal{D}_i)$ where $\theta$ is the parameter vector and $\mathcal{D}_i$ is the data associated with device $i$.

\paragraph{Notation (FAIR):}\label{not:name} Under \fair, we denote the server parameter-vector as $\theta \in R^n$. Each device $i$ has a capacity $0 < \alpha_i \leq 1$ and stores a parameter vector $\psi_i \in R^{m_i}$ where $ m_i \leq \alpha_i n$. Each device is associated with a subspace of dimension $m_i$, which is denoted by the column space of an orthogonal (column) matrix $S_i: n \times m_i$. The optimization functions of device $i$ are denoted as $F'_i(\psi_i, \mathcal{D}_i) = F_i(S_i \psi_i, \mathcal{D}_i)$. We denote $\theta_i = S_i \psi_i$. $\theta_i$ and $\psi_i$ are the same vectors in different bases. We call the basis of $\theta$ the server-basis and that of $\psi$ the device-basis.

\subsection{Hashing-based random projections and associated subspaces}
In this section, we elaborate on how computationally cheap hash functions can be used to create random projection matrices (and hence subspaces), which can be stored in $O(1)$ memory and applied in $O(n)$ time. We also define the reduction and recovery operations for transitioning between the server model and device models.

Consider a client with capacity $\alpha$. \fair{} assigns a subspace to each client based on its capacity. The subspace is defined by a projection matrix, say $S\in R^{\hspace{1pt}n \times m}$ where $m \leq \lfloor \alpha n \rfloor$ Using dense projections such as Gaussian are expensive, both in terms of the memory required to store the projection and to compute the projections of a given vector. Hence, we use hashing-based sparse projections. Examples of good projections for our purpose would include count-min-sketch (CMS) \cite{cormode2005improved} count-sketch (CS) \cite{charikar2002finding}, ROBE \cite{robe1}, ROAST\cite{roast} etc. We will discuss CMS for its simplicity in this section. However, the discussion can be trivially extended to other mappings. In our experiments, we use ROAST mapping. CMS mapping is a binary matrix $S_{cms}$, and is defined as follows:
\begin{equation}
    S_{cms}[i,j] = 1 \iff h(i) = j  \label{eq:cms}
\end{equation}
Note that each row of the matrix has exactly $1$ non-zero and the location is determined by hash function $h:\{0, \dots ,n{-}1\} \rightarrow \{0, \dots , m- 1\}$ where $h$ is drawn from the family of universal hash functions \cite{carter1977universal} that require $O(1)$ memory and $O(1)$ computation. The matrix $S$ is never realized in memory. We only store $h$ to compute $S[i,j]$ on the fly as required using Equation \ref{eq:cms}.

\begin{algorithm*}[t]
\setstretch{1.1}
\SetKwInOut{Parameter}{Requires}
\caption{FAIR: Heterogeneous Federated Averaging in Random Subspaces (Server)}\label{alg:algo1}	
\Parameter{ $N$ devices with data $\{\mathcal{D}_i\}_{i=1}^N$; capacity ratios $\{\alpha_i\}^N_{i=1}$; $0<\alpha_i \leq 1$; loss function $F(\theta, \mathcal{D})$; initial parameter vector at the server, $\theta$.}
\nl \textup{Initialize} $\theta$ 

\nl \For{\textup{round,} $t\in[0,\dots,T - 1]$} { 
        \nl $\theta' \gets 0$\; 
        
         \nl \textup{Generate seed for hash function} $h$\; 
         
         \nl $C \gets$ \textup{a random set of clients}\; 
         
         \nl \textup{Communicate seed to clients in} $C$\; 
         
         \nl $\forall i \in C, S_i \gets $ \textup{the projection matrix corresponding to capacity $\alpha_i$ according to Equation \ref{eq:matrixcut}}\; 
        \tcc{These matrices are not realized, hashing is performed on-the-fly}  
         \nl $\psi_i \gets$ \textup{projection}$(S_i, \theta)$ \textup{according to the Equation $\ref{eq:reduction}$  operation}\; 
         
         \nl \ForEach{\textup{client,} $i \in C$, \textup{in parallel}} {
                 \nl \textup{Send} $\psi_i$\; 
                 
                 \nl $\psi_i' \gets $ \textup{wait for client update}\; 
                 
                 \nl $\theta_i' \gets S_i\psi_i'$ \textup{projection according  to Equation $\ref{eq:recovery}$ operation}\;  
                 
                 \nl $\theta' \gets \theta' + p_i\theta_i'$ \tcc*{$p_i$ are the aggregation weights in fed-avg aggregation is atomic to avoid data race}
         }
     \nl $\theta \gets \theta'$\;
}
\end{algorithm*}
\begin{algorithm*}[t]
\SetKwInOut{Parameter}{Requires}
\caption{FAIR: Heterogeneous Federated Averaging in Random Subspaces (Client $i$)}	
\Parameter{ Data $\{\mathcal{D}_i\}$; capacity ratio $0<\alpha_i \leq 1$; loss function $F(\theta, \mathcal{D})$.}
\vspace{6pt}

\nl \textup{Receive seed for hash function $h$}\;

\nl \textup{Receive $\psi_i$}\;

\nl \textup{Optimize $F_i(\mathcal{D}_i, \psi_i) = F(\mathcal{D}_i, S_i\psi_i)$ for $E$ epochs on local data $\mathcal{D}_i$}\; \tcc{Training done in a memory efficient manner using kernels from \cite{robe1}, \cite{desai2022efficient} which use the \ref{eq:partial} equation}

\nl \textup{Send updated $\psi_i$ to server}\;

\end{algorithm*}

\paragraph{Reduction (server model to client model):} Projecting $\theta$ onto the subspace defined by $S$ to obtain $\psi_S$, is equivalent to finding a solution to the equation:
\begin{equation}
    \psi_S = \textrm{argmin}_{\psi_S} || S \psi_{S} - \theta ||_2  \label{eq:1}
\end{equation}
Generally, this linear system of equations can be solved using normal equations or SGD. However, with a sparse $S$ such as those defined by CMS, we have a simple solution:
\begin{equation}
    \psi_{S_{cms}}[j] = \frac{\sum_{i} \theta[i] \mathbf{I}(h(i){=}j)}{\sum_{i} \mathbf{I}(h(i){=}j)} = S_{cms}^\top\theta / S_{cms}^\top \mathbf{1}^{(n)} \label{eq:reduction}
\end{equation}
This can be implemented by computing $S_{cms}^\top\theta$ and performing element wise division by the counts $S_{cms}^\top \mathbf{1}^{(n)}$ where $\mathbf{1}^{(n)}$ is the $n$-dimensional column vector of $1$s. Due to the sparse $S_{cms}$, this is a $O(n)$ time operation.

\paragraph{Recovery (client model to server model):} The vector $\psi_S \in R^{\hspace{1pt} m}$ is in the basis of $S$. In order to recover the vector in the original $R^n$ space, we compute:
\begin{equation}
    \theta_S = S \psi_S  \label{eq:recovery}
\end{equation}
Note that $\theta_S$ is the vector in the column space of $S$, and $S$ is simply a change of basis matrix.

In typical federated learning settings, the server broadcasts $\theta$ at the beginning of each round, and clients return $\psi_S$ at the end of each round. The reduction operation is required when the server needs to send $\psi_i$, and the recovery operation is required when the server receives $\psi_S$ as shown in figure \ref{fig:fair-process} (right). Alternatively the server can broadcast $\theta$ and the client can project it in a streaming fashion to obtain $\psi_i$. For the discussion, we assume the former setup.

\subsection{Running model on the client without realizing full memory}
As described above, $\psi_S$ is the model representation used on a client device. Under FAIR, the optimization function computation is still $F(S\psi, \mathcal{D})$. In other words, the computation has not changed. How can the model run on the device without incurring $|S\psi| = n$ memory? The idea is straightforward---the client only realizes parts of the full model that are necessary at the time of computation. For instance, in an embedding table lookup, we only need access to some particular embeddings in a batch. See illustration in Figure~\ref{fig:embedding_fair}. To retrieve the embedding parameters located at, say $\theta[a:b]$, we can use:
\begin{equation}
    \theta[a:b] = S[a:b,:]  \psi_S  \label{eq:partial}
\end{equation}
Recent advances in parameter-sharing methods provide efficient kernels based on partial recovery, such as those for embedding tables \cite{robe1} and matrix multiplication \cite{desai2022efficient}.

\subsection{Consistent and collapsible subspaces} \label{consistent-collapsible}
The choice of subspaces is important for effective collaboration of information among devices. As an example, consider two randomly drawn $1$-dimensional subspaces in $R^n$. For a large $n$, with high probability, these subspaces are orthogonal. Thus, the information transferred from one subspace to another via projection is nil as any vector in one subspace, when projected onto another, will go to $\mathbf{0}$. Thus, we must choose a set of consistent and collapsible subspaces as defined below.

\begin{definition}[\textbf{Consistent and Collapsible Subspaces}]
\vspace{3pt} A given set of subspaces defined by columns of matrices $\{S_i\}_{i=1}^N$ where $S_i : n \times m_i$ matrix, is consistent if, for every pair $i,j$, if $m_i = m_j$ then column-space of $S_i$ is the same as the column-space of $S_j$. Also, if $m_i < m_j$, then $\textrm{column-space}(S_i) \subset \textrm{column-space}(S_j)$.
\end{definition}

The compatibility between two subspaces for information sharing can be measured by the angles between the two subspaces. By considering subspaces contained inside one another, we maximize the compatibility of the subspaces. In \fair{}, each unique subspace is identified by a single projection matrix $S$. Thus, a consistent and collapsible subspace associated with all devices, including the server, implies using the exact same basis for all devices that can host the full model. This also implies that if all participating devices can host full models, \fair{} is identical to standard FEDAVG.

We now show how to construct these subspaces using lightweight hash functions when the parameter sizes on devices $\{m_i\}$ are powers of 2. Depending on the capacity $\alpha$, the parameter size, $m_i$, is taken to be the greatest power of $2$ smaller than $\alpha_i n$.

\paragraph{Constructing \textit{consistent} and \textit{collapsible} subspaces using hashing:} Let $m = \max(\{m_i\})$. We draw the hash function $h:\{ 0,\dots,n-1\} \rightarrow \{0,\dots,m-1\}$ from a family of cheap hash functions such as the universal family \cite{carter1977universal}. The set of subspaces is then defined as follows. Let each device have a subspace represented by an orthogonal matrix $S_i$. We define the binary $S_i: n \times m_i$ as follows:
\begin{equation}
    S_i[a,b] = 1  \quad \textrm{iff} \quad h(a) \% m_i = b   \label{eq:matrixcut}
\end{equation}

Note that each row of $S_i$ only has one non-zero element. Thus, each column of $S_i$ is orthogonal to every other column. We can convert this matrix into an orthonormal matrix by normalizing each column to the unit norm.

\begin{theorem}
The set of subspaces $\{S_i\}_{i=1}^N$ of dimensions $\{m_i\}_{i=1}^N$, where each $m_i$ is a power of 2, created using the hash function $h$ as per Equation \ref{eq:matrixcut}, is a consistent and collapsible set of subspaces.
\end{theorem}

This computation is illustrated in Figure \ref{fig:marixfolding} (left). Essentially, we first create a matrix $S$, which is of dimension $n \times m$. Then, in order to create the matrix $S_i$ for subspace $i$, we cut the matrix in columns of width $m/m_i$ and add the pieces together. This also provides an intuition for why this set of subspaces is collapsible: given any two matrices $S_i$ and $S_j$, if $m_i < m_j$, then each column of $S_i$ is the sum of $m_j/m_i$ columns of $S_j$, and hence, column-space($S_i$) $\subset$ column-space($S_j$). We want to emphasize that these computations occur on the fly, and no matrix is actually realized.
%\apd{transitions among models of two compression}

\subsection{FAIR Algorithm}
The complete algorithm of \fair{} is provided in Algorithm~\ref{alg:algo1}. At each round, the server projects the parameter vector $\theta$ onto the subspaces associated with each client sampled for that round using \ref{eq:reduction} operations and sends the parameter in the device-basis, $\{ \psi_i\}$, to the clients. Client $i$ optimizes the function $F(S_i\psi_i, \mathcal{D}_i)$ using local data $\mathcal{D}_i$ with $\psi_i$ residing inside the subspace associated with $S_i$. Clients employ efficient parameter-sharing kernels using \ref{eq:partial} equations, which ensure that the memory footprint of the model remains $|\psi|$. After $E$ steps, the client sends the updated $\psi_i$ back to the server, which is transformed into a server-basis using a \ref{eq:recovery} operation and aggregated with updated models from other clients using standard FEDAVG weighted aggregation.

\ssection{Convergence of FAIR} \label{sec:theory}

\begin{figure*}
    \begin{subfigure}{0.48\textwidth}
    \centering
    \includegraphics[trim={0 0 0 1.35cm},clip,scale=0.45]{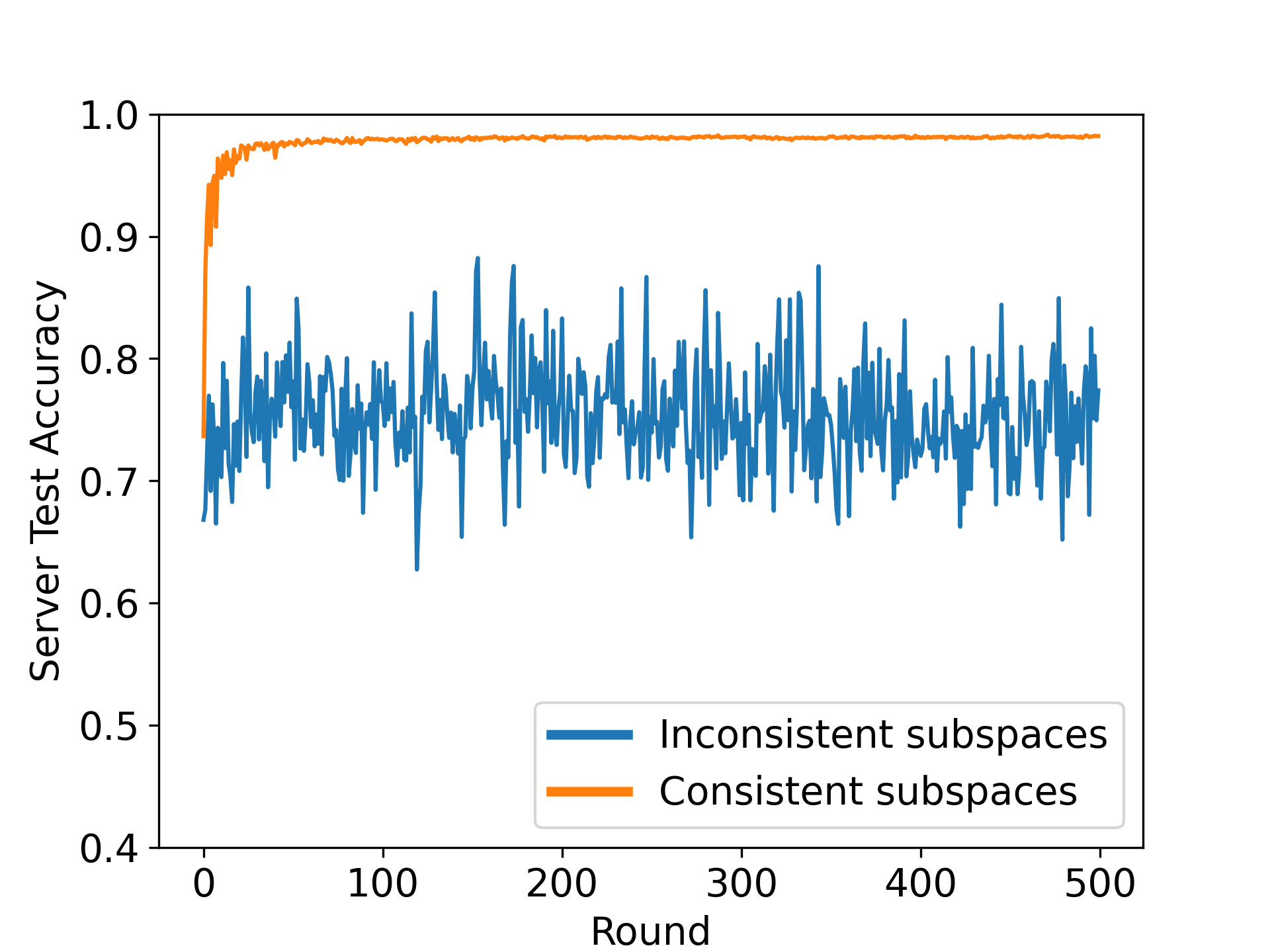}
    \caption{Test accuracy evolution for MNIST classification}
    \end{subfigure} 
    \hfill
    \begin{subfigure}{0.48\textwidth}
    \centering
    \includegraphics[scale=0.45]{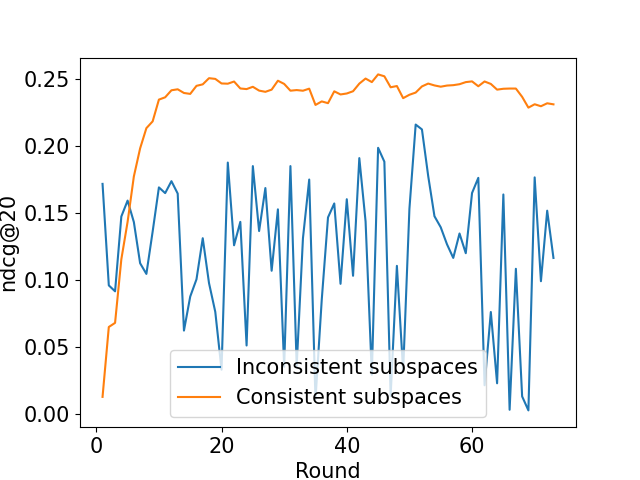}
    \caption{Test ndcg@20 evolution for Goodreads-100 ranking task.}
    \end{subfigure}
    \caption{Global test metrics vs. round for consistent and inconsistent (independently drawn) sub-spaces for MLPs on MNIST (left) and Embedding on Goodreads-100 ranking task (right) trained using \fair{}}\label{fig:subspaces} 
    
\end{figure*}

Empirically, we can see that consistent subspaces are necessary for convergence, as shown in Figure \ref{fig:subspaces}. In this section, we prove the convergence of \fair{} in homogeneous settings where models are compressed equally on all devices. 

Recently, $O(\frac{1}{T})$ convergence of federated learning on non-iid data with full and partial participation of devices was proven \cite{li2019convergence} when the following conditions (paraphrased from \cite{li2019convergence}) are met: (1) $F_1$, ... $F_N$ are L-smooth; (2) $F_1$, .... $F_N$ are $\mu$-strongly convex; (3) variance of stochastic gradients on all the devices is bounded. If $\zeta_t^k$ is the stochastic sample for device $k$ at round $t$, then  $\mathbb{E} || \nabla F_K(\theta^k_t, \zeta_T^k) - \nabla F_K(\theta^k_t, \mathcal{D}_k) ||^2 \leq \sigma_k^2$; and (4) expected squared norm of the stochastic gradients is uniformly bounded, $\mathbb{E}(|| \nabla F_k(\theta^k_t, \zeta^k_t ||^2 \leq G^2)$ for all devices and rounds. For the exact expression of convergence, we refer the reader to \cite{li2019convergence}. The convergence of \fair{} under homogeneous conditions with full participation (alt. partial participation) can be stated as the corollary of Theorem 1 (alt. Theorem 2, 3 in \cite{li2019convergence}). We first need two lemmas to complete the proof.

\begin{lemma}\label{lemma:assumption}
Under \fair{} notation, if $F_1, F_2, \dots, F_N$ satisfy the assumptions 1-4 above, and $F'_1, F'_2, \dots , F'_N$ are:
\begin{equation}
    F'_i(\psi) = F_i(\mathcal{S}\psi)  \qquad \forall i \in \{1,\dots ,N\}
\end{equation}
If $\mathcal{S}$ is an orthonormal matrix, then $F'_1, F'_2,\dots, F'_N$ also satisfy the assumptions 1-4 with the same constants $L, \mu, \sigma_k, G$ for all $k\in \{1,\dots,N\}$.
\end{lemma}

\begin{lemma}\label{lemma:server}
Under \fair{} notation and with a unique subspace matrix $S$ across all the devices, the server parameter-vector $\theta_S$ can always be represented as $\theta_S = S \psi$ for some $\psi$. And the parameter aggregation equation can be written as $\psi = \sum_{i=1}^N  p_i \psi_i$
where $p_i$ are weights of aggregation as per standard $\mathrm{FEDAVG}$.
\end{lemma}
Proof of Lemma \ref{lemma:assumption} is provided in the Appendix. 
With Lemmas~\ref{lemma:assumption} and \ref{lemma:server}, we can cast the entire federated learning with \name under homogeneous settings as federated learning with parameters in the subspace $S$ (even at the server). Thus, the convergence of \name can be stated as a corollary to Theorem 1, 2, 3 from \cite{li2019convergence}. We only state the concise form of Theorem 1 here:

\begin{corollary}
If the functions $F_1, \dots, F_N$ satisfy assumptions 1-4, then under the homogeneous setting of \name (equal dimension of random subspace), the algorithm converges at a rate $O(\frac{1}{T})$.
\end{corollary}

\ssection{Experiments}
\begin{table*}

\centering
\caption{[NDCG@20 (higher is better) on Goodreads-100, AmazonProduct-100]. \fairhet{} has superior quality over \fairhom{}  and \fedhm{} baselines across different capacity schemes. N/A implies that capacity scheme cannot be run. std-deviation ($< 0.005$ for Goodreads-100 and $< 0.009$ for AmazonProduct-100) }
\label{tab:implicit_table_1}
%\resizebox{\linewidth}{!}{
\begin{tabular}{|c|ccc|ccc|}
\hline
                  & \multicolumn{3}{c|}{\textbf{Goodreads-100}}                                                      & \multicolumn{3}{c|}{\textbf{AmazonProducts-100}}                                                 \\ \hline
Central Training  & \multicolumn{3}{c|}{0.28}                                                                        & \multicolumn{3}{c|}{0.20}                                                                        \\ \hline
FEDAVG            & \multicolumn{3}{c|}{0.278}                                                                       & \multicolumn{3}{c|}{0.178}                                                                       \\ \hline
\textbf{Capacity} & \multicolumn{1}{c|}{\textbf{\fairhet{}}} & \multicolumn{1}{c|}{\textbf{\fairhom{}}} & \textbf{\fedhm{}} & \multicolumn{1}{c|}{\textbf{\fairhet{}}} & \multicolumn{1}{c|}{\textbf{\fairhom{}}} & \textbf{\fedhm{}} \\ \hline
1x-2x             & \multicolumn{1}{c|}{0.276}             & \multicolumn{1}{c|}{0.269}             & 0.274          & \multicolumn{1}{c|}{0.13}              & \multicolumn{1}{c|}{0.10}              & 0.078          \\ \hline
1x-4x             & \multicolumn{1}{c|}{0.276}             & \multicolumn{1}{c|}{0.254}             & 0.275          & \multicolumn{1}{c|}{0.14}              & \multicolumn{1}{c|}{0.098}             & 0.041          \\ \hline
1x-8x             & \multicolumn{1}{c|}{0.274}             & \multicolumn{1}{c|}{0.246}             & N/A            & \multicolumn{1}{c|}{0.087}             & \multicolumn{1}{c|}{0.07}              & N/A            \\ \hline
1x-16x            & \multicolumn{1}{c|}{0.276}             & \multicolumn{1}{c|}{0.208}             & N/A            & \multicolumn{1}{c|}{0.056}             & \multicolumn{1}{c|}{0.06}              & N/A            \\ \hline
\end{tabular}
%}
\end{table*}
\begin{table*}
\centering
\caption{NDCG for Goodreads-100 and AmazonProducts-100. Comparison of \fairhet{} with the situation where data from lower-capacity devices is not used (\fulltrunc{}). The columns define percentage of users that can host full models (1x capacity) and others are assumed to have 2x compression. The users are randomly chosen with seed 101. Thus the sets for each column can be non-overlapping.}
\label{tab:data-loss}
\begin{tabular}{|c|cccc|cccc|}
\hline
                    & \multicolumn{4}{c|}{\textbf{Goodreads-100}}                                                                                          & \multicolumn{4}{c|}{\textbf{AmazonProducts-100}}                                                                                             \\ \hline
\textbf{}           & \multicolumn{1}{c|}{\textbf{1\%}} & \multicolumn{1}{c|}{\textbf{20\%}} & \multicolumn{1}{c|}{\textbf{30\%}} & \textbf{50\%} & \multicolumn{1}{c|}{\textbf{1\%}} & \multicolumn{1}{c|}{\textbf{20\%}} & \multicolumn{1}{c|}{\textbf{30\%}} & \textbf{50\%} \\ \hline
\textbf{\fulltrunc{}} & \multicolumn{1}{c|}{0.037}        & \multicolumn{1}{c|}{0.247}         & \multicolumn{1}{c|}{0.238}         & 0.261         & \multicolumn{1}{c|}{0.003}       & \multicolumn{1}{c|}{0.006}         & \multicolumn{1}{c|}{0.05}         & 0.09          \\ \hline
\textbf{\fairhet{}}   & \multicolumn{1}{c|}{0.271}        & \multicolumn{1}{c|}{0.258}         & \multicolumn{1}{c|}{0.27}          & 0.258         & \multicolumn{1}{c|}{0.107}         & \multicolumn{1}{c|}{0.109}          & \multicolumn{1}{c|}{0.110}          & 0.121         \\ \hline
\end{tabular}
%\vspace{-0.25cm}
\end{table*}
\begin{table*}
\centering
\caption{Explicit feedback rating prediction on Goodreads-100 (MSE lower is better). We stop training after 1000 rounds. The \fedhm{} method cannot cater to compression ratios of 32 or beyond for an embedding table of 32 dimension. N/A implies the scheme cannot be run.}
\label{tab:explicit_table_1}
\begin{tabular}{|c|c|c|c|}
\hline
\textbf{Capacity} & \textbf{\fairhet{}} & \textbf{\fairhom{}} & \textbf{\fedhm{}} \\ \hline
8x-16x            & 0.6735            & 0.6891            & 0.7243         \\ \hline
16x-32x           & 0.6716            & 0.6907            & N/A            \\ \hline
32x-64x           & 0.6726            & 0.6916            & N/A            \\ \hline
4x-8x-16x         & 0.6618            & 0.6891            & 0.7206         \\ \hline
8x-16x-32x        & 0.6607            & 0.6907            & N/A            \\ \hline
16x-32x-64x       & 0.6618            & 0.6916            & N/A            \\ \hline
\end{tabular}

\end{table*}
\begin{table*}
\centering
\caption{Test accuracy results of different model capacity settings compared to (uncompressed) full FEDAVG MNIST and FEMNIST datasets. N/A implies the scheme cannot be run.}
%\resizebox{\linewidth}{!}{
\label{tab:general_1}
\begin{tabular}{|c|ccc|ccc|}
\hline
\textbf{}            & \multicolumn{3}{c|}{\textbf{MNIST}}                                                              & \multicolumn{3}{c|}{\textbf{FEMNIST}}                                                            \\ \hline
\textbf{FEDAVG}      & \multicolumn{3}{c|}{0.9841}                                                                      & \multicolumn{3}{c|}{0.7491}                                                                      \\ \hline
\textbf{Capacity}    & \multicolumn{1}{c|}{\textbf{\fairhet{}}} & \multicolumn{1}{c|}{\textbf{\fairhom{}}} & \textbf{\fedhm{}} & \multicolumn{1}{c|}{\textbf{\fairhet{}}} & \multicolumn{1}{c|}{\textbf{\fairhom{}}} & \textbf{\fedhm{}} \\ \hline
2x-4x                & \multicolumn{1}{c|}{0.9822}            & \multicolumn{1}{c|}{0.9829}            & 0.9838         & \multicolumn{1}{c|}{0.7842}            & \multicolumn{1}{c|}{0.7847}            & 0.7804         \\ \hline
2x-4x-8x             & \multicolumn{1}{c|}{0.9823}            & \multicolumn{1}{c|}{0.9779}            & 0.984          & \multicolumn{1}{c|}{0.7691}            & \multicolumn{1}{c|}{0.779}             & 0.7689         \\ \hline
2x-4x-8x-16x-32x-64x & \multicolumn{1}{c|}{0.9636}            & \multicolumn{1}{c|}{0.962}             & 0.9839         & \multicolumn{1}{c|}{0.7409}            & \multicolumn{1}{c|}{0.7114}            & NA             \\ \hline
\end{tabular}
%}
\end{table*}
% \usepackage{multirow}
\begin{table*}
%\centering
\caption{Server model accuracy on MNIST (E=20, K=10) using consistent and inconsistent sub-spaces.} \vspace{2pt}
\label{tab:consistency}
\centering

\begin{tabular}{|c|c|c|c|c|c|c|}
\hline
\textbf{Sub-spaces} & \textbf{COMP} & \multicolumn{1}{l|}{\textbf{\begin{tabular}[c]{@{}c@{}}\name\\Test Acc\end{tabular}}} & \textbf{\begin{tabular}[c]{@{}c@{}}Round\\@ Best\end{tabular}} & \textbf{COMP} & \textbf{\begin{tabular}[c]{@{}c@{}}\name\\Test Acc\end{tabular}} & \textbf{\begin{tabular}[c]{@{}c@{}}Round\\@ Best\end{tabular}} \\ 

\hline

\multirow{3}{*}{\textbf{\begin{tabular}[c]{@{}c@{}}Consistent\end{tabular}}} & $4\times$ & 0.9833 & 472 & $2\times$-$4\times$ & 0.9833 & 461 

\\ \cline{2-7} 

& $8\times$ & 0.981 & 298 & $2\times$-$4\times$-$8\times$ & 0.9815 & 486  

\\ \cline{2-7} 

& $64\times$ & 0.9625 & 234 & $2\times$-$4\times$-$8\times$-$16\times$-$32\times$-$64\times$ & 0.9657 & 192  

\\ \hline

\multirow{3}{*}{\textbf{\begin{tabular}[c]{@{}c@{}}Inconsistent\end{tabular}}} & $4\times$ & 0.8823 & 154 & $2\times$-$4\times$ & 0.8965 & 154

\\ \cline{2-7} 

& $8\times$ & 0.8733 & 154 & $2\times$-$4\times$-$8\times$ & 0.8846 & 154          

\\ \cline{2-7} 

& $64\times$ & 0.7465 & 148 & $2\times$-$4\times$-$8\times$-$16\times$-$32\times$-$64\times$ & 0.8712 & 196               

\\ \hline

\end{tabular}
\end{table*}

To empirically evaluate  \fair{} on CF, we choose both the implicit feedback-based ranking and explicit feedback-based rating prediction tasks. Additionally, we show that \fair{} can be used for general model components by showing some simple general model training under \fair{}. All experiments were performed on Quatro-RTX-8000 GPUs.

\paragraph{Capacity-scheme:} The Ax-Bx capacity notation implies that the first half of users ordered by IDs use Ax compression, and the remaining half use Bx compression. If using $c$ capacities ($A_1$x-$A_2$x-\dots -$A_c$x) we divide users into $c$ groups and use one capacity for each group.

\paragraph{Baselines:} We consider \heterofl{} \cite{diao2020heterofl} and \fedhm{}\cite{yao2021fedhm} from the heterogeneous federated learning literature. We find that both methods are essentially the same in regards to the model (when applied to embedding tables in the first layer, see Appendix \ref{appendix:baselines}). \fedhm{} is more recent and has shown superior performance over \heterofl{} on general federated learning tasks. We thus choose \fedhm{} as the baseline in our experiments. We also define two more baselines. (1) \fairhom{}: This baseline simulates \textbf{"model loss"} i.e., it chooses the smallest capacity as the capacity across all devices and uses \fair{} methodology of communication. In this section, we will refer to \fair{}, which uses heterogeneous capacities as \fairhet{}. (2) \fulltrunc{} : This baseline simulates \textbf{"data loss"} where the full model is trained only using devices that can host the full model (i.e., has 1x capacity).

\subsection{Implicit feedback based ranking CF}
\paragraph{Experimental Setup:}
We choose two datasets\footnote{https://cseweb.ucsd.edu/~jmcauley/datasets.html} (1) Goodreads \cite{wan2018item} and (2) AmazonProduct \cite{mcauley2015image, he2016ups}. We choose the $100$ users with the most interactions and create truncated versions denoted as Goodreads-100 and AmazonProduct-100. We use the NeuMF \cite{he2017neural} model architecture with $8$ factors and $1$ layer MLP. Each client stores two user embeddings and two item embedding tables suitably compressed to its capacity. We use \textit{Bayesian Personalized Ranking} (BPR) loss and evaluate ranking using \textit{Normalized Discounted Cumulative Gain} (ndcg@20) over all items not present in the training set. We check the quality of the server model over the entire test dataset. We evaluate the model every ten rounds. Hyperparameter details are presented in Table~\ref{tab:implhypers} in the Appendix.

\paragraph{Results} The results from the experiment have been tabulated in two tables - Table~\ref{tab:implicit_table_1} and Table~\ref{tab:data-loss}. We make the following observations,
\begin{itemize}
    \item The empirical results show that \fairhet{} provides better quality models than all baselines, irrespective of the capacity scheme. It is evident from the results that \fedhm{} is not suitable for devices that require higher compression, and in such cases \fairhet{} enables training including all devices.
    
    \item \textbf{Model loss:} The model-loss experiment is presented as part of table \ref{tab:implicit_table_1}. The improvement of \fairhet{} over \fairhom{} is significant. This indicates that with \fairhet{}, when we deploy larger models on higher capacity devices and learn richer models, this information is maintained in communication with other devices as well, elevating the overall quality of the trained model. Thus, it is important to use variable-sized models and utilize the varying device capacities to the fullest.
    \item \textbf{Data loss:} The data loss experiment is presented in table \ref{tab:data-loss}. The \fulltrunc{} paradigm, which drops data from devices that cannot host the full model, is impacted significantly, indicating the magnitude of information loss due to excluding data. In contrast, by creating compressed models on these low-capacity devices, we can retain the data and thus maintain the quality.

\end{itemize}

\subsection{Explicit feedback-based rating prediction CF}

\paragraph{Dataset and Model} We choose Goodreads-100 for the rating prediction task. We use the NeuMF-MLP \cite{he2017neural} model with an embedding dimension of $32$ and a $3$ layer MLP. Each client stores one user embedding vector and an item embedding table suitably compressed to its capacity. We use \textit{Mean Square Error} (MSE) loss for both optimization and evaluation.

\textbf{Results:} The results from explicit rating prediction as presented in table \ref{tab:explicit_table_1}. The observations are similar to those from the implicit ranking experiment. \fairhet{} avoids model-loss and data-loss and gives the best performance. Other methods, such as \fedhm{}, cannot cater to low-capacity devices, and even on devices where it can be used, the quality preservation is better with \fairhet{}. More settings and details are included in Appendix \ref{appendix:explicit}.

\subsection{FAIR on general model components}
While \fair{} is best suited for embedding tables, in this section, we explore whether \fair{} can be applied to other model components such as MLP and convolutions since the \fair{} algorithm is general. 

\paragraph{Dataset and Model:}
For MLP, we test a three-layer MLP (784-512-10) on the MNIST \cite{lecun1998mnist} dataset. For the FEMNIST dataset, we test the LeNet\cite{lecun2015lenet} architecture. Data is generated using non-i.i.d Dirichlet distribution \cite{caldas2018leaf}. We use E=20 epochs, T=200, K=10, and 30 devices for both the datasets.

\paragraph{Results:} The results are tabulated in Table~\ref{tab:general_1}.  In capacity schemes that can be run with \fedhm{}, \fair{} is competitive with \fedhm{}, with the latter being slightly better at times. 
However, importantly, when we go to higher compressions, \fedhm{} cannot be used, and in such cases, \fairhet{} still gives us a way to include such low-capacity devices.

\subsection{Ablation: Inconsistent vs. Consistent subspaces}
One of the key ideas in \fair{} is that of consistent and collapsible subspaces. In this section, we show that it is indeed necessary to have these special subspaces for effective collaboration of information and, thus, learning. The results are presented in figure \ref{fig:subspaces} and table \ref{tab:consistency} for various settings. It is clear that with inconsistent subspaces, learning is severely impacted, and thus, learning fails to happen effectively. In contrast, with the correct choice of subspaces - \textit{consistent and collapsible} we can the server model can learn from all the different devices in both homogenous and heterogenous settings.

\ssection{Conclusion}
This paper discusses an important and unexplored problem of learning embedding tables in federated collaborative filtering under heterogeneous capacities. Due to the large size of embedding tables, not all devices can participate in federated learning of entire models. Naive approaches to this problem include training a model using only devices that can host the full model or reducing the model size to include more devices. \fair{} proposes to include all devices with on-device model sizes tuned to their capacities and provides a seamless collaboration of a heterogeneous representation of information. The key idea is using random projections, which have the ability to provide arbitrary compression and \emph{consistent} and \emph{collapsible} subspaces, which make information collaboration possible.
\bibliography{example_paper}
\bibliographystyle{mlsys2024}
\clearpage 
\appendix
\onecolumn
\section{Societal Impact} This work enables federated learning of recommendation systems in hopes of preserving data privacy and providing users with useful recommendation systems that are essential to navigate the vasts amount of content on WWW and other specific platforms. To the best of our knowledge, there is not negative societal impact of our work.

\section{Hyperparameters for implicit ranking experiments.}

\textbf{Hyperparameter search:}
We search for the hyperparameters from the following choices: learning rate $(1e{-}1, 1e{-}2, 1e{-}3, 1e{-}4, 1e{-}5)$, batch size ($256, 512, 1024$), regularization weight for BPR loss ($1e{-}2, 1e{-}4, 1e{-}6$), dropout for MLPs ($0, 0.1, 0.25, 0.5$), number of local epochs in federated learning E($1, 2$). Hyperparameters are chosen by running federated setup for $50$  rounds and central setup for $20$ epochs. We find that most sensitive parameter is the learning rate which is different for different settings while effect of other parameters is not that significant,
we eventually use the following parameter usage :
\begin{table}[h]
\centering
\caption{Hyperparameters used for implicit dataset}
\label{tab:implhypers}
\begin{tabular}{|c|c|c|}
\hline
              & \textbf{Goodreads-100}             & \textbf{AmazonProduct-100}         \\ \hline
E             & 5                                  & 1                                  \\ \hline
num\_neg      & 1                                  & 1                                  \\ \hline
BPR decay     & 1e-6                               & 1e-6                               \\ \hline
MLP dropout   & 0                                  & 0                                  \\ \hline
batchsize     & 512                                & 512                                \\ \hline
learning rate & Best of \{1, 1e-1,1e-2,1e-3,1e-4\} & Best of \{1, 1e-1,1e-2,1e-3,1e-4\} \\ \hline
\end{tabular}
\end{table}

\section{Results to prove convergence}
Proving that the constrained solution search in a subspace also has convergence guarantees as the modified functions have the same guarantees as the assumptions in \cite{li2019convergence}.

In this section, we prove the convergence of \name in a homogeneous setting. We will be leveraging the proof given by \cite{li2019convergence}.

We make the same assumptions on the functions:

\textbf{Assumption 1:} (from \cite{li2019convergence}) $F_1(x), F_2(x), .. F_N(x)$ are all L-smooth for all $\mathbf{v}$ and $\mathbf{w}$, 

Under homogeneous settings, we optimize inside a subspace that is defined by $S$ and use change of variables. The conversion between the two variables is as follows,

\begin{equation}
    x = Sy
\end{equation}
\begin{equation}
    y = (S^\top S)^\dagger S^\top x
\end{equation}

hence, actually we are optimizing for the functions 
$F'_1(y), F'_2(y), .. F'_N(y)$
where 
\begin{equation}
    F'_i(y) = F_i(Sy)
\end{equation}

\paragraph{Claim : } $F'_i(y)$ is also lipschitz smooth.

\begin{align*}
    || \nabla_y F'(y_1) - \nabla_y F'(y_2) || \\
    = || \nabla_y F(Sy_1) - \nabla_y F(Sy_2) || \\
    = || S^\top (\nabla_x F(x_1) - \nabla_x F(x_2) ) ||
\end{align*}
where $x_i = Sy_i$ and using chain rule.

\begin{align*}
    || S^\top (\nabla_x F(x_1) - \nabla_x F(x_2) ) || \\
    \leq ||S^\top|| \; ||(\nabla_x F(x_1) - \nabla_x F(x_2) )|| \\
    = \lambda_{\max} \; ||(\nabla_x F(x_1) - \nabla_x F(x_2) )||
\end{align*}
where $\lambda_{\max}$ is max right-eigen value of the matrix S. Thus, 

\begin{align*}
    || S^\top (\nabla_x F(x_1) - \nabla_x F(x_2) ) || \\
    \leq ||S^\top|| \; ||(\nabla_x F(x_1) - \nabla_x F(x_2) )|| \\
    = \lambda_{\max} L || x_1 - x_2 || \\
    = \lambda_{\max} L || S y_1 - S y_2 ||
    \leq \lambda_{\max}^2 L || y_1 - y_2 ||
\end{align*}

Thus the functions $F_i'(y)$ are also lipschitz smooth with constant $\lambda_{\max}^2L$

\textbf{Assumption 2.}(from \cite{li2019convergence}

$F_1, ... F_N$ are $\mu$-strongly convex.

\begin{align}
    \langle \nabla_y F'(y_1) - \nabla_y F'(y_2) , y_1 - y_2 \rangle \\
    = \langle \nabla_y F(Sy_1) - \nabla_y F(Sy_2) , y_1 - y_2 \rangle \\
    = \langle S^\top \left(\nabla_x F(x_1) - \nabla_x F(x_2) \right) , y_1 - y_2 \rangle \\
    = \langle \left(\nabla_x F(x_1) - \nabla_x F(x_2) \right) , S(y_1 - y_2) \rangle \\
    = \langle \left(\nabla_x F(x_1) - \nabla_x F(x_2) \right) , S(y_1 - y_2) \rangle \\
    = \langle \left(\nabla_x F(x_1) - \nabla_x F(x_2) \right) , x_1 - x_2) \rangle \\
     \geq  \mu ||x_1 -x_2||^2\\
     \geq  \mu ||S (y_1 -y_2)||^2\\
     \geq  \mu \lambda^2_{\min} ||(y_1 -y_2)||^2
 \end{align}
 where $\lambda_{\min}$ is the minimum right eigen value of $S$. Thus, the functions we are optimizing are also strongly convex.

\textbf{Assumption 3.} (from \cite{li2019convergence})

\begin{align}
    \mathbb{E} || \nabla_y F'(y_t, \hat{D}) - \nabla_y F'(y_t,D)||^2 \\
    = \mathbb{E} || \nabla_y F(Sy_t, \hat{D}) - \nabla_y F(Sy_t,D)||^2 \\
    = \mathbb{E} || S^\top (\nabla_x F(x_t, \hat{D}) - \nabla_x F(x_t,D)) ||^2 \\
    \leq \lambda^2_{\max} \mathbb{E} 
     ||(\nabla_x F(x_t,  \hat{D}) - \nabla_x F(x_t,D)) ||^2 \\
     \leq \lambda^2_{\max} \sigma \\
\end{align}

\textbf{Assumption 4} (from \cite{li2019convergence})
stochastic gradients are uniformly bounded. 
\begin{align*}
    E || \nabla_y F'(y, \hat{D}) ||^2 \\
    = E || \nabla_y F(Sy, \hat{D}) ||^2 \\
    = E || S^\top \nabla_x F(x, \hat{D}) ||^2 \\
    \leq  \lambda^2_{\max} E || \nabla_x F(x, \hat{D}) ||^2 \\
    \leq \lambda_{\max}^2 G^2
\end{align*}

If $\mathcal{S}$ is an orthonormal matrix, then $\lambda_{\max} = \lambda_{\min} = 1$. Thus the exact constants work for the modified functions as well. Hence the model converges.

\section{Baselines applied to embedding tables} \label{appendix:baselines}
The HETEROFL and FEDHM applied to embedding tables looks like
\begin{figure}[ht]
    \centering
    \includegraphics[scale=0.15]{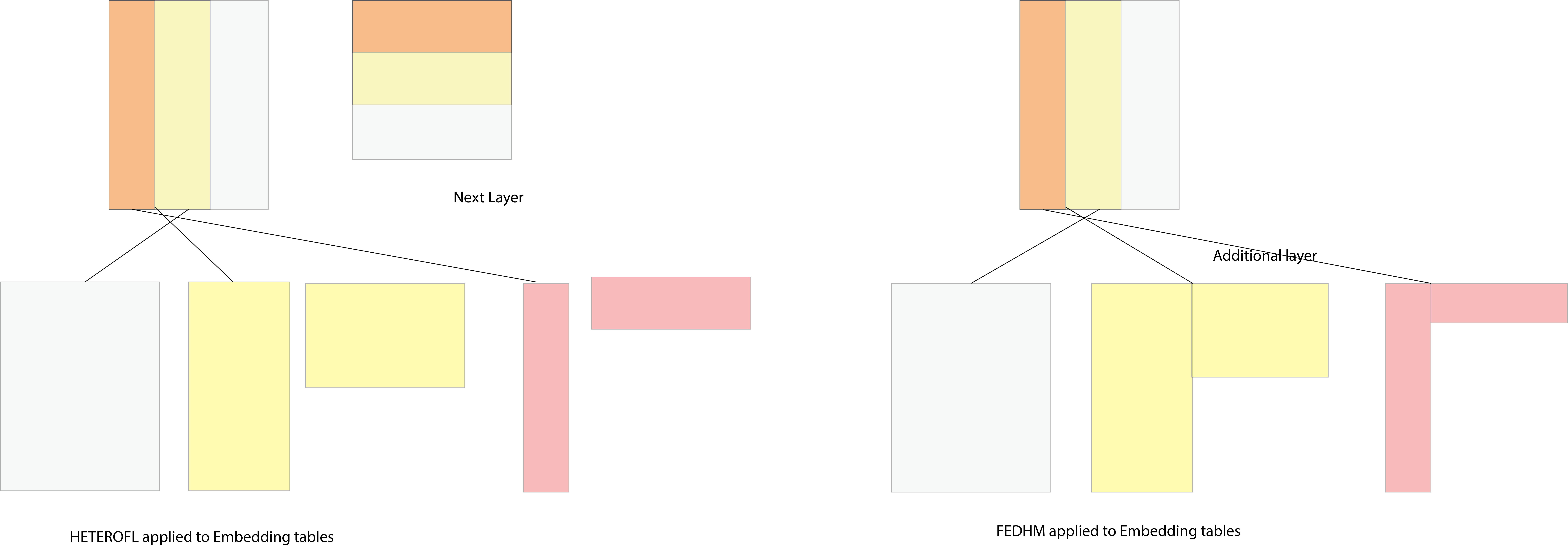}
    \caption{Baselines when applied to embedding tables. Note that the HETEROFL can only be applied to embedding dimension as we need to store some representation of item.  Based on device capacity, one can select reduced dimension for HETEROFL and rank in FEDHM}
    \label{fig:my_label}
\end{figure}

Actually model-wise HETEROFL and FEDHM, when applied to embedding tables, are equivalent. For an embedding table of size $n \times d$ which is reduced to say $n \times k$, then the next layer is modified from $d \times d'$ to $k \times d'$. Thus in some sense it is the low-rank decomposition of the computation of embedding table followed by matrix multiplication. There are some differences in the way aggregation of update would work under HETEROFL and FEDHM. However, for the sake of comparison with FAIR, we consider FEDHM as the all-encompassing method as it is recent and beats HETEROFL on the standard federated tasks.

\section{Additional settings }\label{appendix:explicit}
We present this as a comprehensive presentation of results on Goodreads-100 and MNIST and FEMNIST results. This is an independent section and has a lot of overlap on the data provided in the main paper.
\begin{table}[ht]
\centering
\caption{\textbf{Goodreads-S: Results of FEDAVG (E=1, T=1000, K=10) on two-tower model} with embedding tables for user and items, embeddings are concatenated and passed through a MLP 10-10-1. Model is optimized to reduce MSE for rating prediction. The training is run for 1000 rounds and test mean-square error is reported. \vspace{4pt}}
\label{tab:baseline}
\begin{tabular}{|c|c|c|}
\hline
\textbf{\begin{tabular}[c]{@{}c@{}}Embedding\\ dimension\end{tabular}} & \textbf{Test MSE} & \textbf{\begin{tabular}[c]{@{}c@{}}Round\\ @ Best\end{tabular}} \\ \hline
2                            & 0.8201            & 1000                  \\ \hline
4                            & 0.7464            & 1000                  \\ \hline
8                            & 0.7277            & 1000                  \\ \hline
16                           & 0.6494            & 980                   \\ \hline
32                           & 0.6527            & 920                   \\ \hline
\end{tabular}
\end{table}
\begin{table*}
\centering

\caption{\textbf{Server model quality on Goodreads-S with \name and LowRank (E=1, T=1000, K=10):} results of different compression settings for embedding table storage on devices. The COMP label notation ``$M\times$" denotes a compression ratio of $M$ (e.g., ``$2\times$" implies the embedding table is compressed to 50\% of its original size). Additionally, ``$M\times$" implies that all 100 devices have a compression ratio of M. ``$M_1\times$-$M_2\times$" implies that the devices are separated into two equally-sized groups, with one group having a compression ratio of $M_1$ and the other having a compression ratio of $M_2$. Similarly, ``$M_1\times$-$M_2\times$-$M_3\times$" implies that the devices are separated into three (almost) equally-sized groups, with devices in the three groups having compression ratios of $M_1$, $M_2$, and $M_3$, respectively. We stop training after 1000 rounds. The LowRank method cannot cater to compression ratios of 32 or beyond for an embedding table of 32 dimension \vspace{4pt}}

\label{tab:server-fair}
\resizebox{\linewidth}{!}{
\begin{tabular}{|c|c|c|c|c|c|c|c|c|}
\hline
\textbf{COMP} & \textbf{\begin{tabular}[c]{@{}c@{}}\name\\ Test MSE\end{tabular}} & \textbf{\begin{tabular}[c]{@{}c@{}}Round\\  @ Best\end{tabular}} & \textbf{COMP} & \textbf{\begin{tabular}[c]{@{}c@{}}\name\\ Test MSE\end{tabular}} & \textbf{\begin{tabular}[c]{@{}c@{}}Round \\ @ Best\end{tabular}} & \textbf{COMP}         & \textbf{\begin{tabular}[c]{@{}c@{}}\name\\ Test MSE\end{tabular}} & \textbf{\begin{tabular}[c]{@{}c@{}}Round \\ @ Best\end{tabular}} \\ \hline
$2\times$           & 0.6637                                                                           & 980                                                              & $2\times$-$4\times$        & 0.6601                                                                           & 980                                                              & $2\times$-$4\times$-$8\times$             & 0.6598                                                                           & 980                                                              \\ \hline
$4\times$           & 0.6701                                                                           & 920                                                              & $4\times$-$8\times$        & 0.6659                                                                           & 920                                                              & $4\times$-$8\times$-$16\times$            & 0.6641                                                                           & 980                                                              \\ \hline
$8\times$           & 0.6832                                                                           & 920                                                              & $8\times$-$16\times$       & 0.6735                                                                           & 780                                                              & $8\times$-$16\times$-$32\times$           & 0.6607                                                                           & 740                                                              \\ \hline
$16\times$          & 0.6891                                                                           & 740                                                              & $16\times$-$32\times$      & 0.6716                                                                           & 720                                                              & $16\times$-$32\times$-6$4\times$          & 0.6618                                                                           & 620                                                              \\ \hline
$32\times$          & 0.6907                                                                           & 680                                                              & $32\times$-6$4\times$      & 0.6726                                                                           & 620                                                              &                      &                                                                                  &                                                                  \\ \hline
6$4\times$          & 0.6916                                                                           & 620                                                              &              &                                                                                  &                                                                  & $2\times$-$4\times$-$8\times$-$16\times$-$32\times$-6$4\times$ & 0.6516                                                                           & 780                                                              \\ \hline
\end{tabular}
}
\resizebox{\linewidth}{!}{
\begin{tabular}{|c|c|c|c|c|c|c|c|c|}
\hline
\textbf{COMP} & \textbf{\begin{tabular}[c]{@{}c@{}}LowRank\\ Test MSE\end{tabular}} & \textbf{\begin{tabular}[c]{@{}c@{}}Round \\ @Best\end{tabular}} & \textbf{COMP} & \textbf{\begin{tabular}[c]{@{}c@{}}LowRank\\ Test MSE\end{tabular}} & \textbf{\begin{tabular}[c]{@{}c@{}}Round\\  @Best\end{tabular}} & \textbf{COMP} & \textbf{\begin{tabular}[c]{@{}c@{}}LowRank\\ Test MSE\end{tabular}} & \textbf{\begin{tabular}[c]{@{}c@{}}Round \\ @Best\end{tabular}} \\ \hline
$16\times$          & 0.7092                                                              & 480                                                             & $8\times$-$16\times$       & 0.7243                                                              & 560                                                             & $4\times$-$8\times$-$16\times$    & 0.7206                                                              & 560                                                             \\ \hline
$32\times$          & NA                                                                  & NA                                                              & $16\times$-$32\times$      & NA                                                                  & NA                                                              & \hspace{26.5pt}$8\times$-$16\times$-$32\times$\hspace{26.5pt}   & NA                                                                  & NA                                                              \\ \hline
\end{tabular}
}

\end{table*}

\end{document}